\documentclass[conference]{IEEEtran}
\IEEEoverridecommandlockouts
\usepackage{cite}
\usepackage{amsthm}
\theoremstyle{definition}

\usepackage{amsmath,amssymb,amsfonts}
\usepackage{threeparttable}
\usepackage{graphicx}
\usepackage{textcomp}
\usepackage{algorithm}
\usepackage{xcolor}
\usepackage{algpseudocode}
\usepackage{enumitem}   
\usepackage{subfigure}
\usepackage{hyperref}
\begin{document}

\title{Imitation Learning for Autonomous Driving: Insights from Real-World Testing\\
}

\author{
\IEEEauthorblockN{Hidayet Ersin Dursun}
\IEEEauthorblockA{\textit{AI and Intelligent Systems Lab.} \\
\textit{Istanbul Technical University}\\
Istanbul, Türkiye\\
dursunh20@itu.edu.tr}
\and
\IEEEauthorblockN{Yusuf Güven}
\IEEEauthorblockA{\textit{AI and Intelligent Systems Lab.} \\
\textit{Istanbul Technical University}\\
Istanbul, Türkiye\\
guven18@itu.edu.tr}
\and
\IEEEauthorblockN{Tufan Kumbasar}
\IEEEauthorblockA{\textit{AI and Intelligent Systems Lab.} \\
\textit{Istanbul Technical University}\\
Istanbul, Türkiye\\
kumbasart@itu.edu.tr}
}

\IEEEpubid{%
  \makebox[\columnwidth]{979-8-3315-1088-6/25/\$31.00 \copyright\
2025 IEEE\hfill}%
  \hspace{\columnsep}\makebox[\columnwidth]{}
}
\maketitle

\begin{abstract}
This work focuses on the design of a deep learning-based autonomous driving system deployed and tested on the real-world MIT Racecar to assess its effectiveness in driving scenarios. The Deep Neural Network (DNN) translates raw image inputs into real-time steering commands in an end-to-end learning fashion, following the imitation learning framework. The key design challenge is to ensure that DNN predictions are accurate and fast enough, at a high sampling frequency, and result in smooth vehicle operation under different operating conditions. In this study, we design and compare various DNNs, to identify the most effective approach for real-time autonomous driving. In designing the DNNs, we adopted an incremental design approach that involved enhancing the model capacity and dataset to address the challenges of real-world driving scenarios. We designed a PD system, CNN, CNN-LSTM, and CNN-NODE, and evaluated their performance on the real-world MIT Racecar. While the PD system handled basic lane following, it struggled with sharp turns and lighting variations. The CNN improved steering but lacked temporal awareness, which the CNN-LSTM addressed as it resulted in smooth driving performance. The CNN-NODE performed similarly to the CNN-LSTM in handling driving dynamics, yet with slightly better driving performance. The findings of this research highlight the importance of iterative design processes in developing robust DNNs for autonomous driving applications. The experimental video is available at \textcolor{blue}{\url{https://www.youtube.com/watch?v=FNNYgU--iaY}}.
\end{abstract}


\section{Introduction}
Autonomous Vehicles (AVs) address key transportation challenges such as road safety and traffic efficiency \cite{singh2019deep}. Typically, they use a modular approach with separate modules for perception, path planning, and control \cite{armaugan2020intelligent, fayyad2020}. Although this design improves verifiability, it may reduce computational efficiency due to redundant computations from isolated module operations \cite{pavel2022}. In contrast, the increasingly popular end-to-end driving system converts raw sensory input into control signals using Deep Neural Networks (DNNs). Although this approach sacrifices some verifiability due to its black-box nature, it provides greater efficiency in handling complex tasks \cite{le2022survey, kebria2019deep, yavas1,yavas2}. 

\subsection{Related Work}
In literature, Imitation Learning (IL) is widely recognized as a viable approach for developing AV systems \cite{ kendall2019learning, cai2020probabilistic, robust_behavioral_cloning,pan2020imitation}. A notable application of IL is the DAVE-2 end-to-end driving model, which utilizes three onboard cameras and has been tested in both simulations and real-world scenarios \cite{bojarski2016end}. Following this successful attempt, Cai et al. \cite{vision-racing-rl} highlighted the use of IL to acquire fundamental driving skills by learning from human demonstrations. In \cite{convlstm-line}, a model that combines the Convolutional Neural Network (CNN) with Long-Short-Term Memory (LSTM) networks has been proposed by using the lane detection and tracking approach.

\subsection{Contribution}

In this study, we focus on developing and deploying DNNs for autonomous vehicles that translate raw image input ($\mathbf{X}(k)$) into real-time steering commands (\( w(k) \)). The key design challenge is ensuring that these predictions are both accurate and fast enough, at a high sampling frequency, to enable smooth vehicle operation. Our objective is to explore and compare various methods for generating steering commands from image input, with the aim of identifying the most effective approach in real-time autonomous driving. 

\begin{figure}[t]
    \centering
\includegraphics[width=0.45\textwidth]{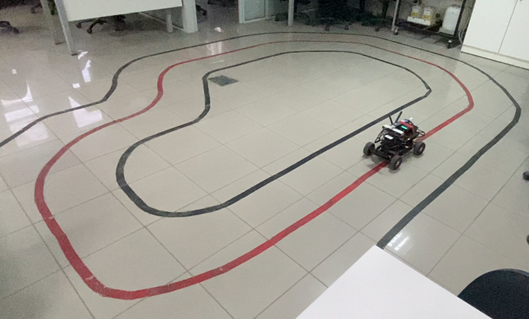}
    \caption{Test environment.}
    \label{fig:test_environment}
\end{figure}

In designing the DNNs, we adopted an incremental design approach that involved enhancing the model capacity and dataset to address the challenges of real-world driving scenarios. We begin by designing a modular steering system for baseline comparison, specifically a classical image processing method combined with a proportional derivative controller (PD) (see Section \ref{exp})). Building on this, we develop DNNs, starting with a CNN to predict steering angles directly from images (see Section \ref{subsubsec:cnn_regression}). To capture the temporal dependencies inherent in driving sequences, we extend the CNN approach by incorporating an LSTM network, which processes time-series data (see Section \ref{subsubsec:cnn_lstm}). Next, we train a CNN integrated with a Neural Ordinary Differential Equation (NODE) model to better capture the evolving dynamics of change in driving (refer to Section \ref{subsubsec:cnn_neural_ode}). Throughout the paper, we assessed the performance of DNNs using the MIT Racecar in a real-world environment, as shown in Fig. \ref{fig:test_environment}. 
We investigated how they handled specific road challenges, such as sharp turns, and evaluated their steering performance at low ($v = 0.6 m/s$) and high speed ($v = 1.2 m/s$). The experimental results presented the efficiency of the DNN models \footnote{\textcolor{blue}{\url{https://www.youtube.com/watch?v=FNNYgU--iaY}}}.

\section{Modular AV Steering Control System} \label{exp}
We first describe the MIT Racecar which will act as a platform to evaluate the IL models. Then, we present modular lane detection and the following system for comparative purposes. 
\subsection{MIT Racecar: Software and Hardware}
\label{subsec:racecar_platform}

The Racecar, developed by MIT, is a research platform designed to test autonomous driving algorithms. It facilitates advancements in autonomous systems and robotics research and offers valuable educational experiences in these fields\cite{karaman2017project, kose}.
% \begin{figure}[htbp]
%     \centering
%     \includegraphics[width=0.25\textwidth]{figures/racecar.png}
%     \caption{MIT Racecar Platform.}
%     \label{fig:racecar_platform}
% \end{figure}
%\subsubsection{Hardware}
The Racecar is built on a RC car chassis, powered by a Brushless DC motor and controlled by an Electronic Speed Controller. It operates with onboard computing powered by the NVIDIA Jetson TX2, which runs real-time AI algorithms. The hardware setup includes three key sensors: a SparkFun 9DoF Razor IMU, a Slamtec RPLidar A2, and a Stereolabs ZED stereo camera that captures 1080p HD video and depth data at 30FPS \cite{stereolabs2023}. A router manages the communication with the ground station.

The Racecar uses the Robot Operating System (ROS) for interfacing and communicating with its onboard systems \cite{quigley2009ros}. ROS offers features such as hardware abstraction, low-level device control, common-use functionality implementation, message transmission across processes, and package management. A ROS system is composed of nodes that interact using a publisher-subscriber architecture, communicating via messages to perform specific tasks or computations in various system components\cite{quigley2009ros}.

\subsection{Modular System: Lane detection and following}
\label{subsec:lane_detection}
The goal is for the Racecar to autonomously navigate using camera input, detecting and tracking a red object representing the lane. A PD controller calculates the steering angle based on the lane's position, maintaining a constant speed.  To achieve this, images are captured at 25 Hz from the left ZED camera using ROS, resized to 640x360 pixels to standardize input and reduce computation, and then a Region of Interest (ROI) is extracted to focus on the relevant areas, like the red line on the ground. The ROI is processed to isolate the target object by converting it to HSV color space, defining a specific color range, and applying a series of image processing techniques to enhance object visibility and reduce noise. 

\begin{figure}[t]
    \centering
    \includegraphics[width=0.45\textwidth]{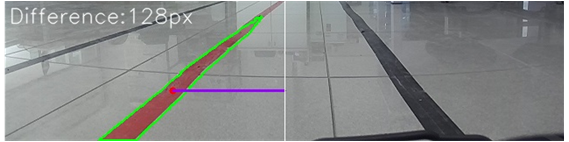}
    \caption{Error representation on an example frame.}
    \label{fig:error_representation}
\end{figure}

The largest contour is identified as the target object for object detection and tracking. The centroid of this contour is calculated and visual cues are added to enhance feedback, as depicted in Fig. ~\ref{fig:error_representation} \cite{bradski2000opencv}. The object's centroid is marked with a red circle, the contour is outlined in green, and the frame's horizontal center is indicated by a white line. Additionally, a purple line extends from the frame center to the object, visually representing the object's position relative to the center.

The horizontal pixel difference between the frame center and the object center, defined as the error (\( e \)), is used as input for the PD controller, which calculates the object's angle from the frame center for real-time tracking. The control law is:
\begin{equation}
w(k) = K_p e(k) + K_d (e(k) - e(k-1))/T_s)
\label{eq}
\end{equation}
where \( T_s \) is the time step (0.04 s), and \( k \) is timestamp. The gain parameters are empirically tuned by testing values between 0.01 and 10 with 0.05 increments. The \( K_p \)  is set to 0.1, and \( K_d \) is set to 0.2. As can be observed from the video, the system performed well at low-speed tests but exhibited some issues over multiple laps, and it was also sensitive to changes in lighting conditions. At high-speed tests, it struggled to track the line accurately, likely due to limitations in processing speed and delays in reacting to rapid changes in the track's curvature.

\section{Imitation Learning for Autonomous Steering}
\label{sec:imitation_learning}
In this section, we focus on how DNNs can be used to convert image input into steering commands for the Racecar. Thus, we aim to learn a DNN model such as:
\begin{equation} \label{dnn}
\text{IL}: f(\mathbf{X}(k)) \rightarrow w(k)   
\end{equation}
where $\mathbf{X}(k)$ represents the RGB frame at the instant $k$ and $w(k)$ the generated steering angle. 

The main challenge here lies in the real-time performance of these models in the Racecar, especially with the changing structure of the test environment track, shown in Fig. ~\ref{fig:test_environment}. To provide a solution, initially, we used a CNN model to imitate the steering behavior of the PD control system. Then, we introduced LSTM layers to capture temporal dependencies, followed by a NODE-based model to improve the car’s ability to handle various driving scenarios\footnote{Github. [Online]. Available: \url{https://github.com/hidayetersindursun/Imitation-Learning-for-Autonomous-Driving-Insights-from-Real-World-Testing}}.

In the real-time implementation of the developed DNN models, we utilized NVIDIA's TensorRT library to optimize the inference process. TensorRT converts the trained model—both its architecture and parameters—into an efficient runtime engine designed for faster and more accurate inference tasks~\cite{tensorrt}. 

\subsection{IL-1: CNN Architecture}
\label{subsubsec:cnn_regression}
\begin{figure}[htbp]
\centerline{\includegraphics[width=0.5\textwidth]{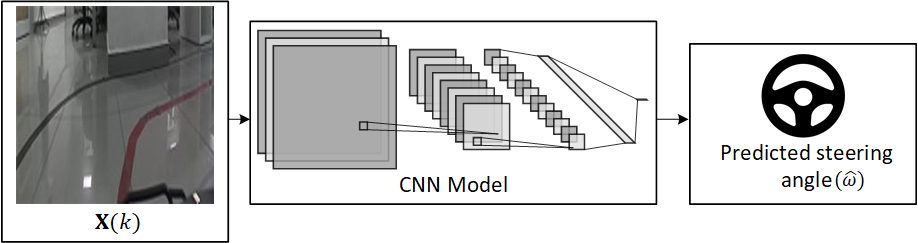}}
\caption{Inference structure of the CNN model.}
\label{fig:inference_structure}
\end{figure}

This section presents the application of IL with CNNs for steering angle prediction ($\hat{w}(k)$) in autonomous driving. The goal is to mimic a modular AV steering system using CNNs. The inference structure is given in Fig. ~\ref{fig:inference_structure}.  

The CNN model was trained using transfer learning with ResNet-18, a well-known model for its deep architecture and extensive pretraining on large datasets~\cite{he2016deep}, modified to suit regression tasks involving steering angle values as targets. The final layers of ResNet-18 were modified by using Matlab Deep Learning Toolbox ~\cite{mathworks2022}, replacing the original fully connected layer with 1000 units followed by a SoftMax layer with smaller fully connected layers of 128, 64, 32, and 1 units. The model was trained using images and the corresponding steering angles collected at 25 Hz using a Jetson TX2, with ROS synchronization ensuring accurate data alignment. The dataset comprises 2000 images for training and 1000 for testing, processed to 224x224 pixels to match ResNet-18's input requirements. Data augmentation introduces lighting variations to enhance the robustness of the dataset. This process resulted in the assembly of a comprehensive training dataset of approximately 6000 images and a testing dataset of approximately 3000 images. For the validation dataset, 15\% of the training images were randomly selected. 
 
\subsubsection{Learning and Testing}
\begin{figure}[t]
\centerline{\includegraphics[width=0.45\textwidth]{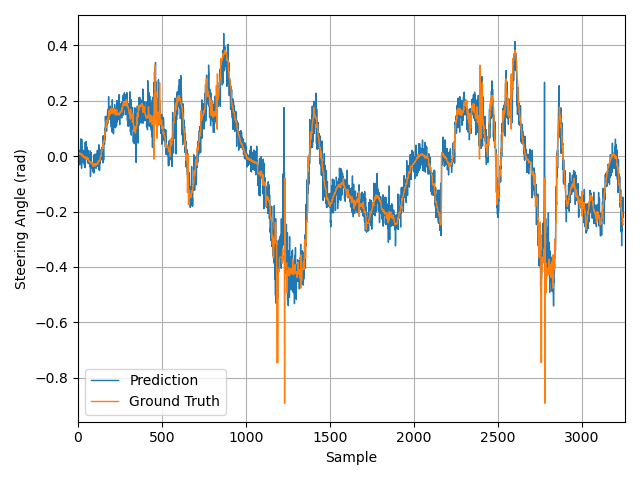}}
\caption{Testing performance of the CNN}
\label{fig:steering_performance}
\end{figure}

The training was performed on Matlab R2022b using an Nvidia GTX 1080Ti GPU. The Adam optimizer was utilized with hyperparameters set to a mini-batch size of 128, a maximum of 15 epochs, and an initial learning rate of 1e-3. Fig. ~\ref{fig:steering_performance} compares predicted and actual steering angles, demonstrating over 90.52\% accuracy of predictions within a 5-degree error margin. 
The testing performance is also evaluated using the Mean Squared Error (MSE) metric, as shown in Table \ref{tab:model_mse}.

\begin{table}[b]
\caption{Testing Performance of DNN models}
\centering
\begin{tabular}{c|ccc}
\hline
        & CNN      & CNN-LSTM & CNN-NODE \\
\hline
MSE& 0.003157   & \textbf{0.001611} & 0.001735 \\
\hline
\end{tabular}
\label{tab:model_mse}
\end{table}

\subsubsection{Real-time Performance Analysis}
During the real-time tests, the model performance varied significantly with speed. At low speed ($v=0.6 m/s$), it performed slightly better than the PD controller, demonstrating generally strong performance but struggling with accuracy over multiple laps, particularly in sharp corners. It was also sensitive to its starting location on the track. This issue arises because the training data do not fully capture all possible road conditions, which hinders the model's ability to adapt to new scenarios on the track. At high speed ($v=1.2 m/s$), the model did not perform effectively, was unable to navigate corners, and exhibited very low performance.

\subsection{IL-2: CNN-LSTM Architecture}
\label{subsubsec:cnn_lstm}

This section presents the IL approach that combines CNNs and LSTMs to improve driving performance. As LSTMs are known to effectively represent~\cite{tuna2022deep}, integrating LSTMs is a natural solution to capture the sequential driving process. Thus, we employ a CNN-LSTM architecture to capture the complex relationship between image inputs and steering angle sequences. CNNs extract image features, which are fed into LSTM layers to learn dynamics within image sequences. This structure enables an understanding of temporal dependencies and improves adaptability to varying driving conditions. 

In designing the CNN-LSTM model, we deviated from the static structure used in \eqref{dnn} and focused on capturing the temporal dynamics of steering. To achieve this, we input sequences of frames  \( \mathbf{X}_e(k) = \{\mathbf{X}(k-2), \mathbf{X}(k-1), \mathbf{X}(k)\} \) into the model follows:  
\begin{equation} \label{lstm}
\text{CNN-LSTM}: f(\mathbf{X}_e(k)) \rightarrow w(k)   
\end{equation}
By using $\mathbf{X_e}(k)$, we aim to mimic how the PD control system or human drivers rely on past visual cues to anticipate future actions. Thus, the temporal context of the time steps \( k-2, k-1, \) and \( k \)  helps the model learn the vehicle’s momentum and speed variations, improving its ability to generate real-time steering commands. Increasing CNN-LSTM capacity by incorporating \( \mathbf{X}_e(k)\) is analogous to how a PD controller responds to both $e(k)$ and its rate of change $(e(k)-e(k-1))$. Just as the proportional term addresses immediate deviations and the derivative term anticipates future changes, the CNN-LSTM leverages consecutive frames to capture temporal dynamics, enhancing the model's ability to adjust steering decisions in real time.
% Temporal dynamics were captured by generating image sequences through the vertical concatenation of consecutive frames at time steps \( k-2, k-1, \) and \( k \). 
% \begin{equation}
% \text{CNN-RNN}: f(X_e(k)) \rightarrow w(k)   
% \end{equation}
% where \( X_e(k) = \{X(k-2), X(k-1), X(k)\} \).
% By integrating these image sequences into the training process, we aim to enhance the model’s ability to predict steering angles with greater accuracy, which in turn can improve the overall speed and performance of the Racecar.

\begin{figure}[b]
\centerline{\includegraphics[width=0.5\textwidth]{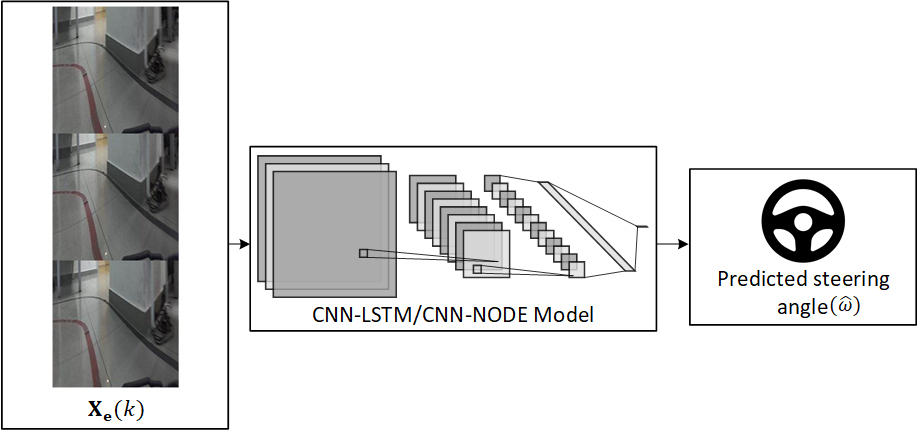}}
\caption{Inference structure of the CNN-LSTM and CNN-NODE.}
\label{fig:inference_structure-cnnlstm}
\end{figure}

The inference structure is given in Fig. ~\ref{fig:inference_structure-cnnlstm}. This structure offers key advantages; it captures temporal dependencies in driving data, enabling smoother and more accurate steering predictions, and enhancing the network's adaptability to varied driving conditions and complex scenarios. However, incorporating LSTMs increases computational complexity, potentially impacting inference performance on the Jetson TX2. To address this, we opted for a different CNN architecture, specifically PilotNet from Nvidia~\cite{bojarski2016end}, which is designed to balance computational efficiency with performance. Additionally, we configured the LSTM layer with 64 units to capture complex patterns in the data and set it to output only the final state to simplify the model's output. We replaced the final layer of the PilotNet with a 64-unit LSTM layer followed by a single fully connected output layer.

To improve model performance in sharp turns, we propose expanding the training dataset to include a wider range of track positions. To support this, we collected personal driving data in the test environment, which consists of images captured during manual driving alongside the corresponding joystick commands used for steering. Our expanded dataset includes 2000 images from the PD controller, as well as an additional 2000 images from the personal driving data we collected. The dataset is further augmented with techniques like brightness adjustment. 

\subsubsection{Learning and Testing}
\begin{figure}[t]
\centerline{\includegraphics[width=0.45\textwidth]{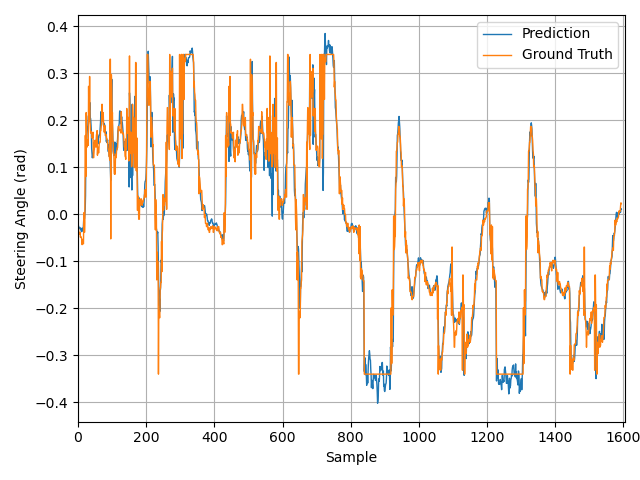}}
\caption{Testing performance of the CNN-LSTM}
\label{fig:test_results}
\end{figure}

We train the CNN-LSTM model within the Matlab Deep Learning environment using the Adam optimizer (learning rate = 1e-3) and train with a mini-batch size of 256 over 100 epochs. The MSE performance of the model on test data is satisfactory as presented in Table \ref{tab:model_mse}. The testing performance of the CNN-LSTM is detailed in Fig. ~\ref{fig:test_results}.
\subsubsection{Real-time Performance Analysis}
\begin{figure}[b]
\centerline{\includegraphics[width=0.45\textwidth]{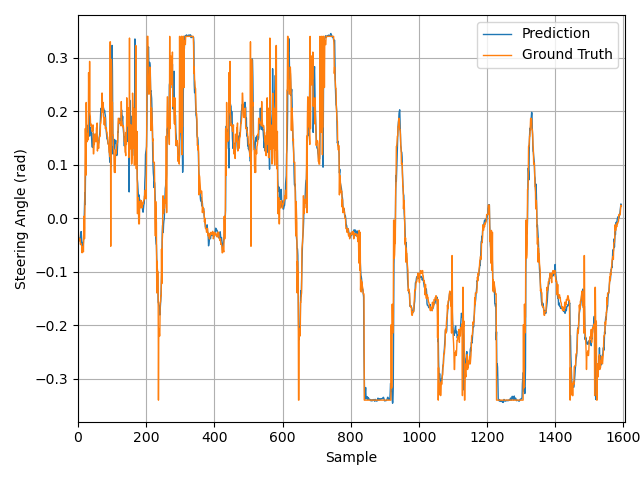}}
\caption{Testing performance of the CNN-LSTM with the enriched training dataset.}
\label{fig:new_test_results}
\end{figure}

The initial model navigated on the test environment smoothly from various starting points but encountered difficulties when handling sharp outer and inner turns. This was due to the limitations in our original dataset, which only included images from the left lens of the ZED camera, capturing outer views during clockwise turns and inner views during counterclockwise turns. As a result, the model struggled with the opposite turns in each direction. To resolve this issue, we expanded the dataset by capturing images from both cameras of the ZED camera, ensuring both views were labeled with the same steering angle. Additionally, we collected two laps of personal driving data in both directions. The new data was merged with the previous dataset, which had been generated using a PD controller. The preprocessing steps included saturating the steering angles, concatenating sequences of three images from both cameras, z-score normalizing the images, and shuffling the sequences. The CNN-LSTM model was retrained using the same hyperparameters, and the test results for this new model are shown in Fig. ~\ref{fig:new_test_results}.
The model exhibited peak real-time performance on the track, successfully navigating all corners and completing over 10 laps. It demonstrated robustness under varied lighting conditions and speeds, performing well up to 1.2 m/s despite data collection at 0.6 m/s. This highlights the LSTM's effectiveness in capturing temporal dependencies, enhancing the model's adaptability. The LSTM integration thus significantly improved performance, reliability, and versatility.

\subsection{IL-3: CNN-NODE Architecture} \label{subsubsec:cnn_neural_ode}

Motivated by the results of CNN-LSTM, we implemented the IL method that integrates NODE which is known to model the continuous-time evolution of neural network states, capturing system behaviors influenced by inputs and states~\cite{ lu2018beyond}. 

In a compact form, the NODE is defined as \cite{chen2018neural}: 
\begin{equation} \label{neural_dynamics}
    \dot{{h}}(t) = f\left({h}(t), t; \theta \right)
\end{equation}
Here, \(f(h(t), t; \theta)\) is a vector field represented by a DNN, that defines the evolution of the hidden state based on its current value \(h(t)\), the time \(t\), and the network's parameters \(\theta\) (which include the weights and biases).
The state of the system at any future time $t_1$ can be obtained by integrating \eqref{neural_dynamics} from initial time $t_0$ to $t_1$ as follows:
\begin{equation} \label{ode_eq}
\begin{split}
{h}(t_1) &= {h}(t_0) + \int_{t_0}^{t_1} f(h(t), t; \theta) \, dt \\
       &= \texttt{ODESolve}(h(t_0), f, t_0, t_1, \theta)\\
\end{split}
\end{equation}
NODEs offer improved memory efficiency and computational cost compared to traditional architectures. They can replace standard residual blocks with an ODE-based approach as shown in \eqref{ode_eq}, achieving competitive performance with fewer parameters and reduced memory usage~\cite{chen2018neural}. Given these advantages, we decided to use NODEs instead of LSTM layers to learn image sequences' temporal features. 

The CNN-NODE structure is identical to the input-output structure of the CNN-LSTM given in \eqref{lstm} and has the same inference structure as shown in Fig~\ref{fig:inference_structure-cnnlstm}. The CNN-NODE architecture starts with PilotNet's CNN layers for spatial feature extraction, followed by a NODE layer implemented using \textit{odeint} function~\cite{torchdiffeq}, replacing the LSTM layers. Fully connected layers (50, 25, 1) complete the model for final predictions. Our dataset remains the same with the CNN-LSTM section.

\subsubsection{Learning and Testing}
The training of CNN-NODE models is implemented in the Pytorch framework using \textit{torchdiffeq} package~\cite{torchdiffeq}. PyTorch offers a more efficient development process for our target NVIDIA Jetson platform, with optimized performance for embedded systems~\cite{pytorch2021}. Similar to the training of the CNN-LSTM, we used the Adam optimizer (learning rate 1e-3, mini-batch size 256) for 100 epochs. To solve the integration step in \eqref{ode_eq}, we defined the Euler solver. 

\begin{figure}[t]
\centerline{\includegraphics[width=0.45\textwidth]{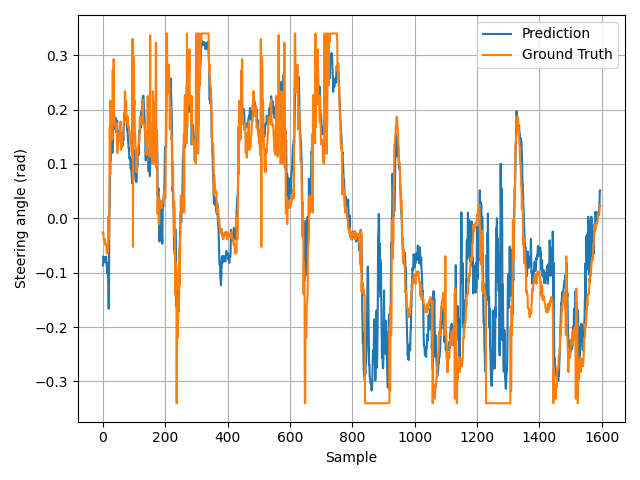}}
\caption{Testing performance of the CNN-NODE with Euler solver.}
\label{fig:test_results_euler}
\end{figure}

\begin{figure}[t]
\centerline{\includegraphics[width=0.45\textwidth]{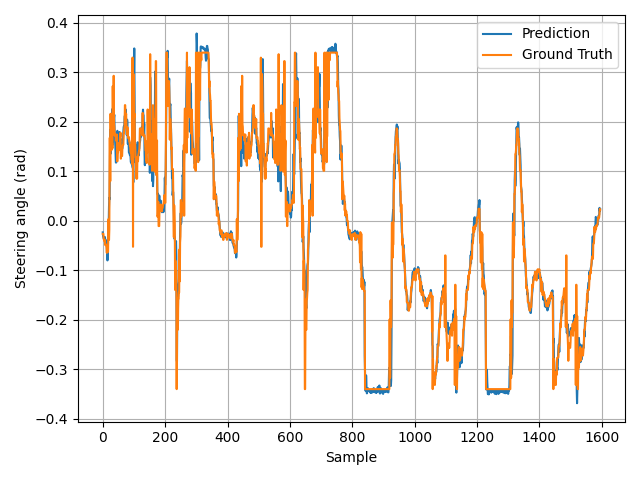}}
\caption{Testing performance of the CNN-NODE with RK4 solver.}
\label{fig:test_results_rk4}
\end{figure}

We tested the CNN-NODE model with the existing testing dataset defined for CNN-LSTM. The test results are shown in Fig. ~\ref{fig:test_results_euler}. The test showed the model underperformed, handling straight lines well but struggling with sharp turns. We suspected the ODE solver choice and switched from Euler to Runge-Kutta fourth-order (RK4). Following the retraining, we tested the CNN-NODE model with the RK4 solver. Fig. ~\ref{fig:test_results_rk4} presents these test results. The model with the RK4 solver outperformed the Euler-based model.


\subsubsection{Real-time Performance Analysis}
The CNN-NODE model handled sharp corners well and remained stable from various starting points. It performed effectively at speeds up to 1.2 m/s, despite initial data collection at 0.6 m/s. This confirmed the NODE model's comparable performance to the CNN-LSTM model, proving its practicality for real-world applications. Also,  the results confirm the RK4 solver's efficacy over the Euler method in real-world applications. 

\section{Conclusion and Future Work}

In this paper, we tackled the steering control problem of AV through IL. In the design of the DNNs, to address the challenges of real-world driving scenarios, we adopted an incremental design approach that involved enhancing the DNN capacity and dataset. From the results, we can summarize that:  
\begin{itemize}
    \item The modular system with PD controller, while effective for basic lane following, struggled with sharp turns and lighting variations, showing limitations in adapting to complex driving scenarios.
    \item The CNN model improved steering angle prediction, but struggled in sharp turns during continuous real-time tests due to its inability to handle temporal dependencies
    \item The  CNN-LSTM model provided a smooth steering and better handling of complex driving situations by incorporating temporal information. It performed well even in challenging conditions such as sharp turns and varying speeds, completing multiple laps without significant drift. 
    \item The CNN-NODE model demonstrated comparable performance to the CNN-LSTM, particularly with the RK4 solver, handling continuous driving dynamics effectively.
\end{itemize}
We can conclude that the presented CNN-LSTM and CNN-NODE were found to be the most robust and reliable for real-time autonomous driving tasks. 

Maybe more importantly, this study provides valuable insights from real-world experiments that should be taken into account during the design and deployment of DNNs for autonomous driving. By strategically modifying both the input and model structure, alongside enriching the training dataset with diverse driving scenarios, we have significantly enhanced the DNN's capacity to adapt to a wide range of driving conditions. This incremental design approach not only improves the robustness and accuracy of the models but also carefully considers practical limitations, such as inference time, ensuring that the DNNs can operate effectively in real-time applications. The findings of this study underscore the importance of iterative design processes in developing reliable autonomous driving systems that can perform optimally under varying environmental challenges.

Future work aims to enhance the robustness of DNNs in dynamic and complex scenarios by utilizing uncertainty quantification methods.

\section*{Acknowledgment}
The authors acknowledge using ChatGPT to refine the grammar and enhance the English language expressions.

\bibliographystyle{IEEEtran}
\bibliography{cites}
\end{document}